# Fast, Accurate Thin-Structure Obstacle Detection for Autonomous Mobile Robots

Chen Zhou *
Peking University
zhouch@pku.edu.cn

Jiaolong Yang
Microsoft Research
jiaoyan@microsoft.com

Chunshui Zhao
Microsoft Research
chunzhao@microsoft.com

Gang Hua
Microsoft Research
ganghua@microsoft.com

## Abstract

*Safety is paramount for mobile robotic platforms such as self-driving cars and unmanned aerial vehicles. This work is devoted to a task that is indispensable for safety yet was largely overlooked in the past – detecting obstacles that are of very thin structures, such as wires, cables and tree branches. This is a challenging problem, as thin objects can be problematic for active sensors such as lidar and sonar and even for stereo cameras. In this work, we propose to use video sequences for thin obstacle detection. We represent obstacles with edges in the video frames, and reconstruct them in 3D using efficient edge-based visual odometry techniques. We provide both a monocular camera solution and a stereo camera solution. The former incorporates Inertial Measurement Unit (IMU) data to solve scale ambiguity, while the latter enjoys a novel, purely vision-based solution. Experiments demonstrated that the proposed methods are fast and able to detect thin obstacles robustly and accurately under various conditions.*

## 1. Introduction

Obstacle detection (OD) is a vital component for autonomous robot maneuver. While the task has been extensively studied in robotics and vision community [31, 28, 22, 42, 36] and several consumer-level products have become available [2, 3, 4], detecting obstacles in less constrained environments with various obstacle types remains a challenging problem. Besides, an OD algorithm should be fast and efficient, such that they can be deployed onto low-power on-board devices or embedded systems, allowing for real-time collision avoidance.

In this paper, we focus ourselves on a scenario that hitherto has received little investigation in the obstacle detection literature, i.e., detecting thin obstacles (e.g. wires, cables or tree branches) in the scene. This task is of practical importance for Unmanned Aerial Vehicles (UAV) such as a drone: crashing into cables / tree branches has become an important reason for UAV accidents. Moreover, thin obstacle detection can also enhance the safety for ground mobile robots such a self-driving car or an indoor robot. For example, it is desired that an indoor moving robot stops in front of a laptop power cord instead of dragging the laptop off the desk. However, detecting these types of obstacles is particularly challenging and presents difficulties to today's OD systems. Active sensors, such as sonar, radar or lidar, provide high accuracy distance measurements but usually suffer from low resolution or high cost. IR structure light (e.g. Kinect) is vulnerable in outdoor scenes. Passive sensors, such as a stereo camera, can provide images with high spatial resolution, but thin obstacles still can be easily missed during stereo matching due to their extremely small image coverage and the background clutter.

To tackle these difficulties, we propose to perform thin obstacle detection using *video sequences* from a monocular camera or a stereo camera pair. More specifically, we represent obstacles with *edges* in the video frames, i.e., image pixels exhibiting large gradients, and focus on recovering their 3D locations. The advantages of adopting an edge-based representation for our application are twofold. First, while thin obstacles like wires or branches are difficult to extract using region or patch based approaches, they can be more easily detected by a proper edge detector. Second, as critical structural information of the scene is also preserved in the edge map, a compact representation is obtained and efficient computation can be carried out, which is critical for embedded systems.

With the edge representation, there are three goals we seek to achieve for our method:

- *Obstacle Identification*: The edges of the obstacle should be extracted, and be complete enough that the obstacle will not be missed.

- *Depth Recovery*: The 3D coordinates of the edges need to be recovered, and be accurate enough that obstacle avoidance actions can be safely performed.

- *Efficient Solution*: The algorithm should be efficient

---

*Work performed while interning at Microsoft Research.

enough to run on-board in real-time with limited computational resources.

These three goals are important for successful thin obstacle detection. While the second and third goals are common for OD systems, we give some remarks for the first one. In our case, the thin lines or cables as obstacles may cross the whole image, and missing a portion of the line may lead to collision. However, for a classical region-based OD system targeting at regularly shaped objects, missing the same portion of an object will probably be acceptable, as long as some margin around the object is reserved. Therefore, the completeness problem is more critical in our case.

Inspired by the works of [25, 27, 16], we propose to use an image edge detector to detect edges in conjunction with edge-based Visual Odometry (VO) techniques to achieve the above goals. Put simply, we detect edge points from images, reconstructing them to 3D using video sequences and efficient VO techniques, and finally identifying obstacles in the front. We present both a monocular camera solution with IMU data incorporation and a stereo camera solution.

As our edge-based method identifies thin obstacles by their recovered 3D positions, there's no need to distinguish between physically thin objects and large objects with color edges: classifying edges belonging to a large object which are close to the robot as obstacles is *correct*. In this sense, though designed to detect thin objects, our method is also capable of detecting generic obstacles with texture edges. Of course, large textureless or transparent objects cannot be handled by our method due to the inherent limitation of vision-based techniques. Nonetheless, our goal is to provide a complementary solution for small and thin objects, and large objects can be easily detected by other approaches such as active sensors. Indeed, we have developed a prototype system combing our method and some low-cost sonar sensors, which is extremely robust and reliable in practice.

In summary, this paper has the following contributions:

- We present fast and accurate solutions to the challenging problem of thin obstacle detection, which has been largely ignored in the OD literature.
- We employed edge-based VO techniques [25, 27, 16] and extend them in various ways such as incorporating IMU information. In particular, we proposed the first edge-based VO algorithm for stereo cameras, to the best of our knowledge.
- We developed a mobile robotic system combining our thin obstacle detection algorithm and sonar sensors, enabling detecting obstacles of general types and avoiding collision reliably.

Our full method runs at 17 fps on an onboard PC of a Turtlebot with a quad-core Celeron 1.5GHz CPU, processing sequences of 640×360 stereo image pairs.

## 2. Related Work

Our system is related to several lines of work in robotics and computer vision community, which we briefly review in this section.

**Obstacle Detection:** Vision-based obstacle detection methods can be generally grouped into monocular or stereo based ones. Monocular methods typically estimate the depth of the obstacles by exploiting supervised learning techniques or monocular depth cues. In [33], an obstacle avoidance system for ground vehicle is proposed. Images are divided into vertical stripes and the depth of each stripe is predicted by a regression model using texture features. The more recent work of [31] exploits a deep encoder-decoder network for depth prediction with image and optical flow as input, however a GPU is needed to run the algorithm. Monocular depth cues have also been extensively explored. Motion parallax is utilized in optical flow based methods [10, 14] where the robot moves in the direction of minor flow amplitude. [36] detects obstacles by monitoring their relative size change across frames. SURF features [8] are extracted from images and template matching is used to detect size change of the features. Saliency detection models have also been used for obstacle detection [28].

Stereo camera setting has the advantage of providing absolute distance estimates to the obstacles. Many of the OD systems for autonomous driving assume the existence of a ground plane [45], and detect obstacles raising from the ground by tessellation of the 2D / 3D space and clustering the disparities from stereo matching [29, 40, 11]. A survey can be found in [9]. Although performing well for autonomous driving, such methods are usually too computationally expensive for embedded systems. For stereo-based OD systems on UAVs, fast computation is often achieved by making some compromise to a complete, global stereo matching method. For example, [7] achieves high frame rate on a mobile-CPU by matching only fixed disparities and fusing these measurements in a push-broom fashion. Others have resorted to FPGA implementations [24, 39].

While the above methods yield robust results for medium-sized and regularly shaped obstacles, little attention has been paid to the detection of thin obstacles such as cables or branches. Recently, Ramos et al. [43, 45] have raised a relevant problem of detecting small unexpected road hazards. The difficulty presented by small object size and unknown object types is addressed by combining deep semantic labeling and Stixel-based geometric modeling [45]. However, their method relies on the assumption that the ground plane exists while we do not, and their region based semantic labeling and geometric modeling strategy is still not suitable for thin obstacle detection as in our case.

**Semi-Dense Visual Odometry (VO) / SLAM:** VO / SLAM algorithms have been widely used for robot localization and position control [6, 21]. However, classical VO / SLAM algorithms that track only sparse features are not particularly suitable for application of identifying obstacles. Fortunately, VO / SLAM algorithms producing semi-dense depth maps which capture the structural information of the scene have become more popular recently, and our system is closely related to these semi-dense methods.

Engel et al. [20] propose a monocular VO algorithm that tracks camera motion with direct image alignment. The inverse depth of each pixel with non-negligible gradient is represented as a Gaussian probability distribution, and updated via a probabilistic Bayesian model. The algorithm is later extended to a key-frame based full SLAM system [18], and further extended to work with stereo cameras [19].

Another line of work exploits edges [25, 27, 16] as features for monocular VO or SLAM. Compared to the *direct* methods above which minimizes photometric error for camera tracking [17], in these works edges are explicitly extracted and matched across frames, and geometric error is minimized during camera tracking.

Both lines of work above produce semi-dense depth maps that could be applied to obstacle detection. We choose the representation with explicit edge extraction in this work for two reasons. First, the explicit edge extraction process gives us more control on detecting thin objects like wires. Second, when similar structural information in the image is captured, an edge map that has gone through edge thinning and cleaning contains fewer edge pixels to process than direct methods, therefore more suitable for embedded systems with limited computing budgets.

**Semi-Dense Stereo Reconstruction:** Recently, Ramalingam et al. [44] and Pillai et al. [41] have proposed methods that perform semi-dense stereo matching at very high frame rates. Beginning with sparse keypoint matching, [44] matches new edge points that satisfy cross-ratio constraints on a local planar region. [41] adopts a coarse to fine strategy, iteratively performing triangulation on the current sparse depth map and re-sampling depth on edge points.

The major issue of these methods is that they implicitly rely on the geometric assumption that the scene is locally planar [31]. Unfortunately, this assumption usually does not hold near the thin obstacles we seek to detect.

## 3. Vision-Based Thin Obstacle Detection

In this work, we are interested in detecting thin obstacles with edge representation. As mentioned in Section 1, there are three goals we seek to achieve for our method: *i)* reliable obstacle identification, *ii)* accurate depth recovery, and *iii)* high efficiency. This section presents our vision-based technique to achieve these goals.

Stated succinctly, we represent obstacles with edges in the consecutive images captured by a monocular or stereo camera, and perform obstacle detection using three main steps: edge extraction, edge 3D reconstruction, and obstacle labeling from edge depth maps.

**Notations:** An edge pixel is represented as a tuple $\mathbf{e} = \{\mathbf{p}, \mathbf{g}, d, \sigma\}$, where $\mathbf{p}$ is the image edge coordinate and $\mathbf{g}$ is the edge gradient. $d$ denotes the *inverse* depth of an edge, and $\sigma$ represents the variance of the inverse depth.

We denote the rotation between the current frame and the new frame as $\mathbf{w} \in so(3)$, and translation as $\mathbf{v} \in \mathbb{R}^3$. $\mathbf{R} = \exp(\mathbf{w}) \in SO(3)$ denotes the rotation matrix. In particular, a 3D point $\mathbf{p}_c$ in the current frame coordinate is transformed to the new frame coordinate as $\mathbf{p}_n = \mathbf{R}\mathbf{p}_c + \mathbf{v}$. We also use a a six dimensional vector $\xi = \{\mathbf{w}, \mathbf{v}\} \in se(3)$ as a compact representation of the Euclidean transformation. $\pi$ denotes the projective function that projects a 3D point in the camera coordinate to image coordinate.

### 3.1. Monocular Thin Obstacle Detection

As a starting point, we present in this section our basic obstacle detection algorithm for the monocular cases, where IMU data can be incorporated to resolve scale ambiguity. Later on we show how the method can be extended to the stereo cases without the need for IMU data.

#### 3.1.1 Edge Extraction

Extracting edges in the images, though seemingly an easy task, is important in our case. It requires a deliberate algorithm design which takes into account various factors including recall / completeness, efficiency, accuracy, etc.

We propose to use a DoG detector [32] for edge detection in combination with a consequent Canny-style hypothesis linking step [13]. The DoG detector is chosen because of its good repetitivity. The hypothesis linking step is introduced to improve the recall of weak edges which are common on thin obstacles. It also allows a more aggressive threshold setting for edge detection: our experiments show that it enables our method to extract $10\% - 20\%$ fewer edges for one image without affecting the thin obstacle detection results, thus improving the efficiency.

Note that we use the regular grid coordinates for the detected edge points, despite subpixel localization can be estimated by parametric fitting (e.g., [30] fits 3D Trivariate quadratics in the DoG pyramid, and [25] fits 2D planes on the image of second derivatives). We empirically found such a subpixel localization step unnecessary in our case: it is time-consuming and has little impact on the depth estimation results.

#### 3.1.2 Edge 3D Reconstruction

After extracting image edges, the next step is to reconstruct these edges to 3D, i.e., estimating their depth.

Our reconstruction method is built upon the recent edge-based VO algorithm proposed by Tarrio et al. [25]. As a light-weight VO system, their algorithm stores only a local *inverse* depth map of the current frame, with the inverse depth of each edge pixel represented as a Gaussian distribution similar to [20]. When a new frame arrives, the camera motion is tracked by fitting the edge map from the current frame to the edge map of the new frame. Specifically, the following geometric error is minimized (see [25, 46]):

$$E_o(\mathbf{w},\mathbf{v}) = \rho((W(\mathbf{p}_i,d_i,\xi) - \bar{\mathbf{p}}_i) \cdot \bar{\mathbf{g}}_i) \quad (1)$$

where $W$ is the warping function that projects an edge pixel $\mathbf{p}_i$ in the current frame into the next frame, taking the image coordinate $\mathbf{p}_i$, inverse depth $d_i$ and the transformation $\xi \in se(3)$ between the two frames. $\bar{\mathbf{p}}_i$ is the corresponding edge in the next frame, found by searching along the gradient direction of the reprojected pixel $\mathbf{p}_i$, and $\bar{\mathbf{g}}_i$ the gradient direction of $\bar{\mathbf{p}}_i$. Mapping is done by epipolar search from the new edge map to the current one. The inverse depth map of the current frame is propagated to the new frame using camera motion, and updated with new observations obtained during the epipolar search. The data is merged in an EKF (Extended Kalman Filter) manner. For more details of the algorithm, we refer the readers to [25].

VO algorithms usually focus on accurately recovering the camera trajectory, whereas for our application, we are concerned more about the completeness of thin obstacle edge detection, and the accuracy of the local inverse depth map. To this end, we extend the original algorithm of [25] by *i)* modeling the uncertainty in edge matching and *ii)* incorporating IMU data, described respectively as follows.

**Modeling Uncertainty in Edge Matching:** A major difficulty for edge-based VO is that edge features are hard to match, both due to the aperture problem and a lack of effective matching descriptors. In the system, this data association problem is approached by projecting the new frame edges to the last frame, and then searching for the closest edge along the epipolar line that satisfies criterions like gradient direction/magnitude consistency and motion consistency [16, 25].

However, the matching criteria for edges are weak. False matchings are common when the initial depth estimation is inaccurate, and multiple similar edges may present in the search range. We address this problem within the EKF fusion framework. To be more concrete, when searching for a match, we collect all the edges that satisfy the criterions in the epipolar search range, and compute their position variance along the epipolar line. For non-ambiguous matching, the variance will be small, whereas when multiple candidates are present, the variance will be large, and subsequently will have a smaller impact on data fusion.

**Incorporating IMU Data** In this work, we use inertial data information primarily for resolving the scale ambiguity for monocular cameras. Some existing works such as [38] consider the VO system as a blackbox and incorporates the scale as a variable in the robot state. Instead, we use the readings from inertial sensors both for initialization and as prior for our vision-based camera tracking procedure.

More specifically, for many mobile robotic platforms, metric 6D robot poses can be estimated from inertial sensors (e.g. using off-the-self libraries such as [34] that integrate sensor readings using EKF) and converted to the camera coordinate system. We denote the relative motion between two consecutive frames obtained in this way as $\mathbf{w}_0 \in so(3)$ and $\mathbf{v}_0 \in \mathbb{R}^3$. Note that $\mathbf{w}_0$ and $\mathbf{v}_0$ are subjected to noise in practice. We modify the objective function to be

$$E(\mathbf{w},\mathbf{v}) = E_o(\mathbf{w},\mathbf{v}) + \lambda_{\mathbf{w}}||\mathbf{w}-\mathbf{w}_0||^2 + \lambda_{\mathbf{v}}||\mathbf{v}-\mathbf{v}_0||^2 \quad (2)$$

where $E_o(\mathbf{w},\mathbf{v})$ denotes the original geometric error in Eq. 1, and the two quadratic terms are priors to regularize the final solution closer to $(\mathbf{w}_0, \mathbf{v}_0)$. We minimize Eq. 2 using the Levenberg-Marquardt algorithm [35] initialized with $(\mathbf{w}_0, \mathbf{v}_0)$, instead of the zero and constant velocity models used in [25] for initialization.

### 3.1.3 Obstacle Labeling on Edge Depth Maps

With the metric depth recovered, we are now ready for thin obstacle identification.

Ideally, back-reprojected edge pixels falling into a predefined 3D volume $\mathbf{S}$ in front of the camera should be labeled as obstacle pixels. However, as in other OD methods [9, 40], the raw depth maps are usually noisy and post-processing is necessary. For robust obstacle labeling, we only consider edge pixels with stable (inverse) depth estimations that have been observed and matched across multiple frames. For each edge $e_i$, besides its image position $\mathbf{p}_i$ and inverse depth $d_i$, we also store its variance $\sigma_i$ and the number of frames $t_i$ it has been successfully matched as criteria for obstacle labeling.

Observing that noisy edges are usually scattered in the depth map, we optionally perform a filtering step on small connected components that have been labeled. For efficiency reason, the filtering step is not performed if the number of initially labeled obstacle edges are below a threshold $cnt_l$ (indicating unlikely existence of any obstacles) or above a threshold $cnt_h$ (indicating highly likely existence of obstacles). The algorithm is summarized in Algorithm 1.

The FILTERSMALLCONNECTEDCOMPONENTS procedure removes small connected components labeled as obstacles. Two obstacle edge pixels with $L_1$-norm distance less than $n_t$ (set as 4 in our experiments) is defined to be neighboring. The procedure is efficiently implemented by searching for connected components on a resized image $I_r$ by a factor of $n_t$. Each pixel of $I_r$ stores the number of obstacle pixels of the corresponding $n_t \times n_t$ block in the original image, and the connected component size of obstacle pixels

**Algorithm 1** Obstacle Edge Pixels Labeling

**Input:** List of edges $e_i = \{\mathbf{p}_i, d_i, \sigma_i, t_i\}$,
Thresholds $\sigma_{th}, t_{th}, cnt_l, cnt_h$, Obstacle Space Volume **S**
**Output:** List of obstacle edgel labelings **O**
$cnt \leftarrow 0$
**for each** edge pixel $e_i$ **do**
    **if** $\sigma_i < \sigma_{th}$ and $t_{th} < t_i$ and $\pi^{-1}(\mathbf{p}_i, d_i) \in \mathbf{S}$ **then**
        $o_i = true$    ▷ The $i$th edgel labeled as obstacle
        $cnt \leftarrow cnt + 1$
**if** $cnt \in [cnt_l, cnt_h]$ **then**
    $\mathbf{O} \leftarrow$ FILTERSMALLCONNECTEDCOMPONENTS($\mathbf{O}$)
**return O**

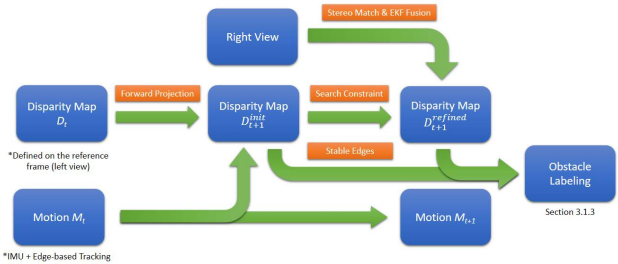

Figure 1: **Workflow of Our Stereo Obstacle Detection**

in the original image is computed by summing up the pixel values of connected components in $I_r$.

## 3.2. Stereo Thin Obstacle Detection

In this section, we extend the monocular OD method to work with a stereo camera, which combines spatial and temporal stereo. It offers a number of benefits compared to a monocular scheme. Apart from resolving the scale ambiguity, matching and reconstruction are now possible for degenerate camera movements, and the necessity of a bootstrapping step is erased. Moreover, the different epipolar line directions for spatial and temporal matching renders them highly complementary: while the matching is ambiguous in one direction, it may be more clear in another.

Our stereo OD method also has three steps as the monocular method, with the edge extraction and depth map based obstacle labeling steps exactly the same (See Section 3.1.1 and 3.1.3 respectively). The edge 3D reconstruction step is based on a novel, edge-based stereo VO algorithm.

In the stereo setting, we still use a Gaussian distribution on the inverse depth (i.e. $d, \sigma$) for each edge pixel. The relationship between stereo disparity $u$ and *inverse* depth $d$ is $u = Bfd$, where $B$ is the baseline length and $f$ is the camera focal length. Similarly, variance in disparity is computed as $\sigma_u = Bf\sigma$. We assume a calibrated stereo camera here thus the input images are rectified. We use the left view as the reference view, and the stereo matching result to refine the inverse depth on the reference view. The workflow is shown in Fig. 1, with some detailed explanations below.

**Inverse Depth Propagation:** Given the current frame $I_t$ and the new frame $I_{t+1}$, camera motion $\xi$ is tracked with IMU as in the monocular case. The inverse depth map $D_t$ (i.e. inverse depth distributions) is propagated to the next frame via the found correspondences during the tracking stage, yielding an initial inverse depth map $D_{t+1}^{init}$ on the reference view of frame $t+1$.

**Edge-based Stereo Matching:** In this step, edge-based stereo matching is performed on the new frame pair. $D_{t+1}^{init}$ is used to constrain the stereo search range. In particular, for an edge pixel with inverse depth distribution $d, \sigma$ in $D_{t+1}^{init}$, we search for corresponding edges only in the range $[u - 2\sigma_u, u + 2\sigma_u]$ in the right view, which is very efficient when the estimation converges. Best matching is found in a similar way as in the monocular case where we perform gradient direction / magnitude and motion consistency check (Section 3.1). Uninitialized edges have large values of $\sigma$, therefore a full search along the epipolar line will be performed.

**EKF Fusion:** The disparities from stereo matching $D_{stereo}$ are treated as observations, and fused with $D_{t+1}^{init}$ again using EKF, yielding the refined inverse depth distribution $D_{t+1}^{refined}$. The temporal matching result is also fused in the same way as before.

**A Remark:** Engel et al. [19] have used a similar scheme to extend their semi-dense LSD-SLAM to the stereo setting. Our approach is different from theirs in that as an *indirect* method, we rely on edge representation for tracking and mapping, whereas theirs is a *direct* method [1]. To the best of our knowledge, our algorithm is the first edge-based stereo VO algorithm.

## 4. An Obstacle Detection System

Having presented our core technique – vision-based thin obstacle detection, in this section we present a system which combines our vision-based OD and ultrasound sonar sensors, arriving at a fully functioning, extremely robust OD system. The sonar sensors are used to detect large textureless or transparent objects, e.g., a large white wall or glass.

The overall system is shown in Fig. 2a. Our prototype system is implemented on a Turtlebot with a Kobuki base. For vision data acquisition, we use a ZED stereo camera [5]. For inertial data, we use the ROS package `robot_pose_ekf` to integrate wheel odometry and gyroscope data and providing robot poses. A hand-eye calibration procedure [47] is carried out to obtain the transform between the camera frame and the robot base.

---
[1] See [17] for a brief review on direct and indirect methods for visual odometry / SLAM

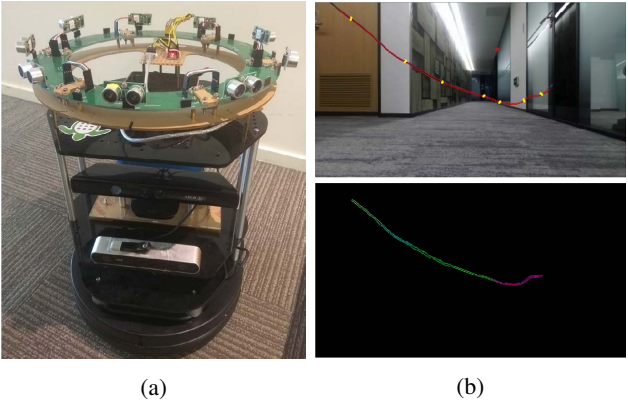

(a)              (b)

Figure 2: **Robot Platform and Ground Truth Labeling.** (a) Our mobile robot platform. Note that the Kinect is not used; we use the ZED camera below (in silver). (b) The top image shows the labeled contour of the thin obstacle, and a series of user-specified corresponding points. The bottom image shows the corresponding depth map.

We installed a sonar array on top of the Turtlebot. Nine sonar sensors are assembled in a circular pattern, clockwise in top view, dividing 360 degrees by constant intervals of 40 degrees. The sonar array returns measurements at 9Hz at 5m range, or 14Hz at 3m range.

Our OD system makes an alert sound and stops when either the vision system detects an obstacle, or the sonar module detects one. The vision module sends out a detection signal if the edges labeled as obstacle is above a threshold, and the sonar module issues a detection if for any of the sonars, the average measurements of the past 5 readings are below a distance threshold. Unlike traditional obstacle avoidance systems that deal with regularly shaped obstacles raising from the ground, the thin obstacles we seek to detect can cross the whole image, and a proper strategy to bypass the obstacle is not obvious. Therefore, we simply force stopping when an obstacle is detected.

## 5. Experiments

This section presents the results of our proposed approach and comparisons to baseline methods. The platform described in Section 4 is used as the testbed. Since the sonar part is well established, our primary focus is to evaluate our vision-based thin obstacle detection algorithms.

### 5.1. Dataset

To our knowledge, there is no existing dataset focusing on evaluating thin obstacle detection. Therefore, we recorded a novel dataset for our purpose, where the Turtlebot was driven in various indoor environments (corridor, meeting room, office, etc.) with different lighting conditions and background clutter. Video sequences were captured by the ZED camera mounted on the robot when the robot was approaching thin objects. Synchronized IMU readings and wheel odometry were also recorded. In the experiments, we use a total of 10 videos captured in six different indoor environments, as shown in Fig. 4.

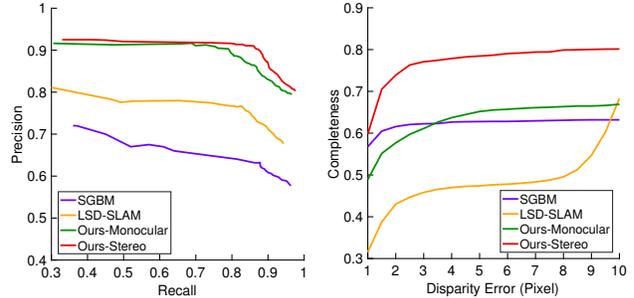

Figure 3: **Quantitative Results.** The left figure shows the precision-recall curve of our detection algorithm, and the right figure shows the completeness measure.

To evaluate the accuracy and completeness of our methods, the ground truth of the thin obstacles is vital. More concretely, we need the extents (edges) as well as the depths of the obstacles to evaluate completeness and accuracy. However, obtaining such data is non-trivial. The resolution of many common sensors turns out to be too low for ground truth capture (e.g. the pre-mounted Kinect sensor on Turtlebot). We address this problem by interactive labeling. The extents of the thin obstacles are sketched out by exploiting the interactive segmentation method of [37] (i.e. Intelligent Scissors). And for depth, the annotator is instructed to roughly specify a series of corresponding points on the thin obstacle, then the exact correspondence is searched considering the epipolar constraint. Finally, the depth in-between is computed by interpolation. A labeled stereo image is shown in Fig. 2b for illustration.

We labeled five frames for each of the 10 videos, resulting in 50 images used for the evaluation. The frames are selected based on proper robot-thin obstacle distances and with some time intervals within each video.

### 5.2. Methods

We evaluate two of our methods: a monocular version with IMU data, and a stereo version. As baseline methods, we choose the Semi-Global Block Matching (SGBM) method [23] from `OpenCV` and the LSD-SLAM work by Engel et al. [18]. The SGBM method itself has been used as ground truth for some semi-dense reconstruction methods [44, 7]. However, here we investigate how it performs with respect to thin objects. LSD-SLAM is a sophisticated state-of-the-art SLAM system exploiting semi-dense representation, which also has the potential to be exploited for thin obstacle detection. As only the monocular version is publicly available, in the experiments we use stereo matching results as initialization to the first several frames, pro-viding initial scene and scale information (we found this greatly improved the quality).

### 5.3. Quantitative Evaluation

Quantitatively, we are interested in measuring the performance of our methods in terms of accuracy and completeness, both of which are very important indexes to ensure safety.

**Accuracy:** To measure the accuracy, we take into consideration both correct prediction for the foreground (i.e. ground truth obstacle) and for the background (which in our case doesn't have ground truth depth). In brief, OD is treated as a classification problem. Let $d$ denote the distance of the current pixel to the closest ground truth (GT) edge, and $s$ be the smallest disparity difference to GT edges, the following quantities are recorded:

| | |
|---|---|
| True Positive (TP) | $d < \tau_0$ and $s < \tau_1$ |
| True Negative (TN) | $d > \tau_0$ and $s > \tau_1$ |
| False Positive (FP) | $d > \tau_0$ and $s < \tau_1$ |
| False Negative (FN) | $d < \tau_0$ and $s > \tau_1$ |

The Precision-Recall (PR) curve in Fig. 3 is generated by fixing $\tau_0$ as 2 pixels and varying $\tau_1$.

It can be seen from Fig. 3 that our methods, both monocular and binocular, have much larger Area Under PR Curve (AUCPR) compared to SGBM and LSD-SLAM methods, indicating significantly better performances. Moreover, as expected, our stereo method slightly outperforms our monocular method.

**Completeness:** We calculate completeness as the percentage of ground truth obstacle edges covered by correct predictions. A ground truth obstacle edge is covered, if a reconstructed edge pixel with disparity difference less than $\tau$ is within 2 pixel's distance. Fig. 3 shows that our stereo method outperforms other methods by wide margins. Our monocular method is only inferior to SGBM when the allowed disparity error is small, which is reasonable because of the higher accuracy achieved by dense stereo matching. The LSD-SLAM has lower completeness not until the allowed disparity error reaches 10.

### 5.4. Qualitative Evaluation

We present some qualitative results in Fig. 4. The readers are suggested to zoom in to see the difference.

For SGBM (2nd column), the margin on the left is caused by the `OpenCV` function. LSD-SLAM is a keyframe based method, non-key frames are only used for tracking and refinement of the depth on the key frames. To obtain the depth map for any frame, we project the point cloud of the latest key frame to the current frame.

Apart from this, there are several remarks we want to make about the results. First, note how by incorporating stereo matching in our method, the noisy edges on the detected obstacles becomes more clear (Row 1), and the error introduced by the regularization averaging process on the prominent vertical edge is reduced (Row 2). Also, as an indirect method, the explicit edge extraction stage helps to better recover the thin line in front of the cluttered background, whereas the LSD-SLAM method mixes the thin line edge pixels and background pixels (Row 1). For Row 4, note the serious "fattening" phenomenon on the earphone wire, which frequently occurs when SGBM is applied to thin objects. The last row shows a case in which combining stereo matching information may not help. The highly non-frontal parallel ground plane with repetitive texture causes the classical SGBM method to fail (2nd column), and similar thing happens for our stereo version. It can be observed that the stereo version yields slightly worse results in the ground part (some edges are estimated too far) than our monocular version. However, our method didn't fail as badly as SGBM in this case, due to the integration of temporal information.

Table 1: **Speed Comparison (in fps).** Measured on a 1.5GHZ quad-core Celeron onboard CPU.

| | SGBM | LSD-SLAM | Ours-Monocular | Ours-Stereo |
|---|---|---|---|---|
| 640×480 | 4.97 | 14.26 | 37.56 | 17.59 |
| 1280×720 | 1.12 | 8.73 | 12.65 | 5.26 |

### 5.5. Running Speed

Finally we present our evaluation of the efficiency. Note that running speed is also important in practice, as higher speed ensures higher safety especially when the robot is in fast motion. Moreover, efficient methods are more favorable for embedded and low-power systems.

Table 1 shows the running speed of different methods on two image sizes. The timing is measured on a 1.5GHZ quad-core Celeron CPU on the Turtlebot. It can be seen that our monocular method is $1.5 \sim 2.5$ times faster that LSD-SLAM, and our stereo method is $4 \sim 5$ times faster than SGBM. For real-time performance or nearly so, we can run on $640 \times 480$ images on the Turtlebot system at over 17fps.

## 6. Thin Obstacle Detection on a Drone

Besides ground robots, we also demonstrate the effectiveness of our system on a drone. We ported our system to a DJI Matrice 100 quadcopter with the Guidance system [1] installed. The Guidance system itself is a sensing system equipped with stereo cameras and sonar sensors, capable of providing stereo image pairs of size $320 \times 240$ at 20Hz via a USB connection. Disparity maps can be obtained at the same frame rate via hardware-based computation. In a preliminary experiment, we compared the disparities computed by our system with the ones provided by the Guidance system. From Fig. 5, it can be clearly seen that our method provides superior thin obstacle detection than the DJI Guidance system.

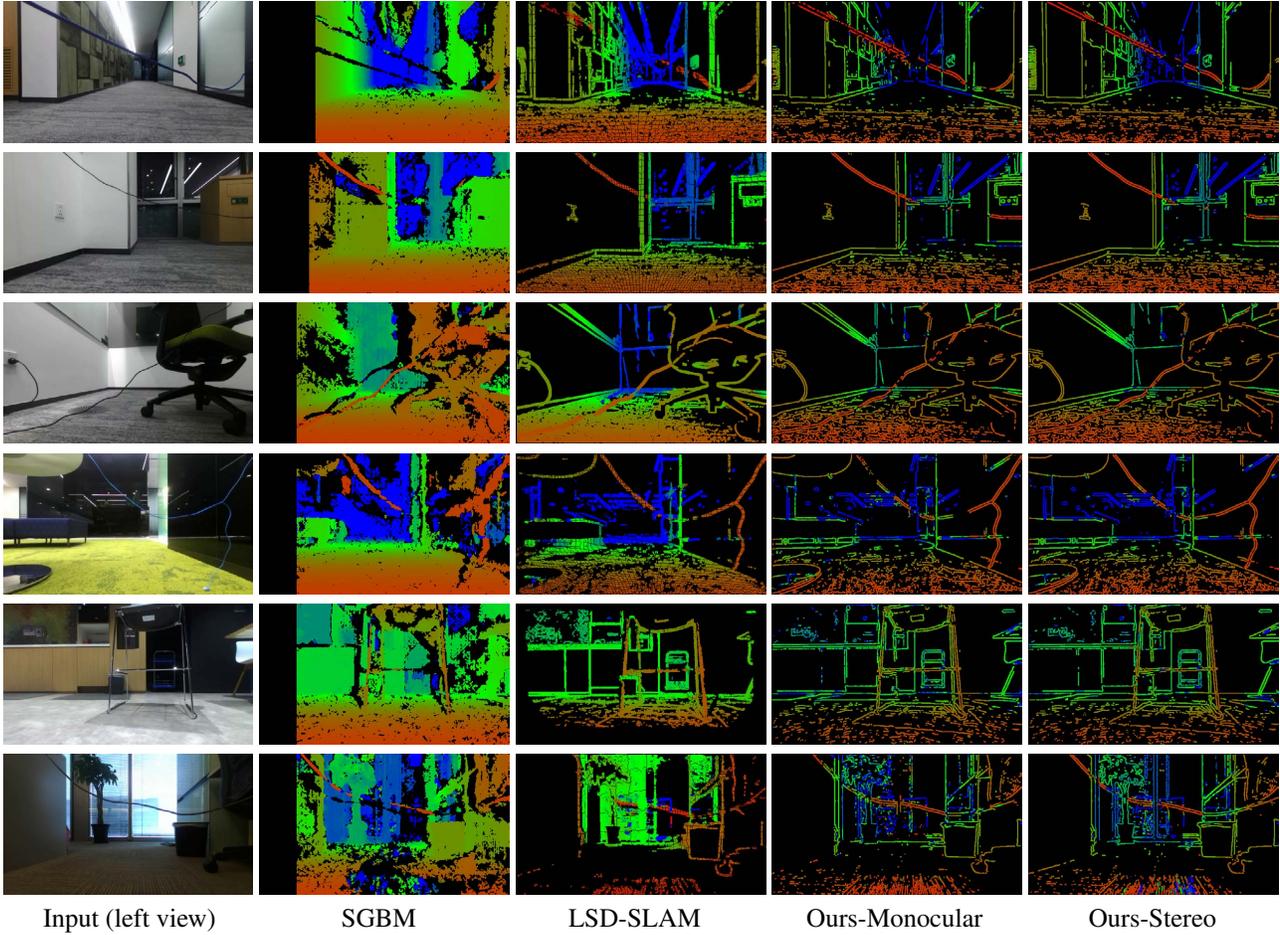

| Input (left view) | SGBM | LSD-SLAM | Ours-Monocular | Ours-Stereo |

Figure 4: **Qualitative Results and Comparison to Baseline Methods.** Best viewed on screen with zoom-in.

## 7. Conclusions and Future Work

In this paper, we studied the thin-structure obstacle detection problem, and presented vision-based methods to tackle this problem. Our method represents obstacles with image edges, and reconstructs them using visual odometry techniques. A monocular camera solution as well as a stereo camera solution are provided, both of which are fast and accurate as shown in the experiments. The system is further extended to a drone and preliminary experiments show encouraging results.

Currently, our vision-based methods are restricted to static scenes. Extending the system to handle dynamic scenes would be important. How our edge-based method can be combined with existing dense stereo matching / visual odometry methods for better performance is also an interesting direction to investigate.

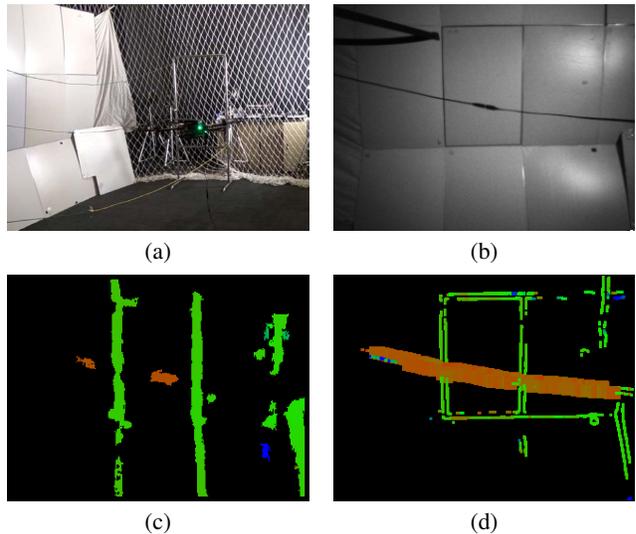

Figure 5: **Thin Obstacle Detection on a Drone.** (a)The scene for experiment. (b) Camera captured image (left view). (c) Disparity map by the DJI Guidance system (noisy regions with area less than 50 pixels are removed). (d) Disparity map by our method. The detected thin obstacles are displayed with a larger line width.